\documentclass{article}
\usepackage[preprint]{spconf}
\usepackage[ruled,linesnumbered]{algorithm2e}
\usepackage{float}
\usepackage{epsfig}
\usepackage{amsmath}
\usepackage{amssymb}
\usepackage{xcolor} 
\usepackage{booktabs}
\usepackage{multirow} 
\usepackage{amsfonts}
\usepackage{caption}
\usepackage{subcaption}
\usepackage{pifont}
\usepackage{makecell}
\usepackage{soul}
\usepackage[accsupp]{axessibility}

\usepackage{tabularx}
\usepackage{stackengine}
\usepackage{bbding}
\usepackage{diagbox} 
\usepackage[hidelinks]{hyperref}

\usepackage{newfloat}
\usepackage{listings}

\let\OLDthebibliography\thebibliography

\renewcommand\thebibliography[1]{
  \OLDthebibliography{#1}
  \setlength{\parskip}{0pt}
  \setlength{\itemsep}{0pt plus 0.3ex}
}

\begin{document}\sloppy

\def\x{{\mathbf x}}
\def\L{{\cal L}}

\title{Exploring Iterative Refinement with Diffusion Models for Video Grounding}
%
\name{\normalsize
    Xiao Liang*\textsuperscript{\rm 1},
    Tao Shi*\textsuperscript{\rm 1},\thanks{\; * These authors contributed equally.}
    Yaoyuan Liang\textsuperscript{\rm 1},
    Te Tao\textsuperscript{\rm 1},
    Shao-Lun Huang\textsuperscript{\rm 1}\thanks{\; $\dagger$ Corresponding author.}
}
\address{\normalsize
    \textsuperscript{\rm 1}Shenzhen International Graduate School, Tsinghua University\\
    \normalsize \{liangx22, shitao21, liang-yy21, taot22\}@mails.tsinghua.edu.cn, 
    shaolun.huang@sz.tsinghua.edu.cn \\
}


\maketitle
\begin{abstract}
\label{sec:abs}

\noindent Video grounding aims to localize the target moment in an untrimmed video corresponding to a given sentence query. 
Existing methods typically select the best prediction from a set of predefined proposals or directly regress the target span in a single-shot manner, resulting in the absence of a systematical and well-crafted prediction refinement process. 
In this paper, we propose DiffusionVG, a novel framework with diffusion models that formulates video grounding as a conditional generative task, where the target span is generated from Gaussian noise inputs and iteratively refined in the reverse diffusion process. 
During training, DiffusionVG progressively adds noise to the target span with a fixed forward diffusion process and learns to recover the target span in the reverse diffusion process. 
In inference, DiffusionVG can effectively generate the target span from Gaussian noise inputs by the learned reverse diffusion process conditioned on the video-sentence representations. 
Without bells and whistles, our DiffusionVG demonstrates superior performance compared to existing state-of-the-art methods on mainstream Charades-STA, ActivityNet Captions, and TACoS benchmarks.

\end{abstract}

\section{Introduction}
\label{sec:intro}

\begin{figure}[t]
\centering
\includegraphics[width=0.99\linewidth]{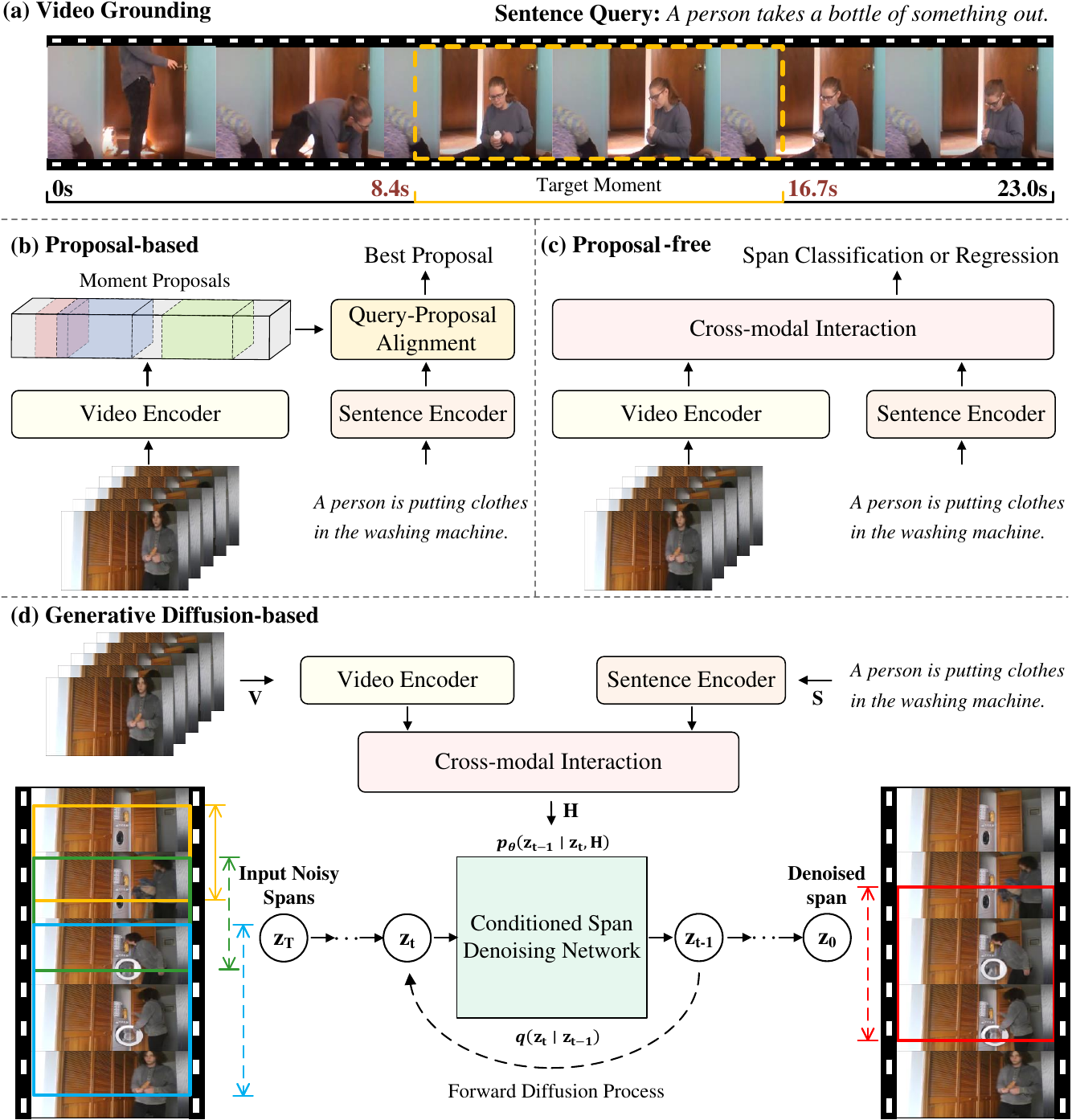}
\caption{(a) An illustration of the video grounding task, which aims to locate a target moment semantically corresponding to a sentence query in a video. (b) Pipeline of proposal-based VG methods. (c) Pipeline of proposal-free VG methods. (d) Our proposed DiffusionVG, which formulates VG as a conditional generative task with diffusion models. 
}
\label{fig:teaser}
\end{figure}

Video grounding (VG) is a fundamental task in video-language understanding, which focuses on identifying the start and end timestamps of a target moment (termed as target span) that exhibits the highest semantic correspondence with a sentence query in an untrimmed video. As shown in Fig~\ref{fig:teaser}(a), this task is challenging since it requires not only separate understanding of the video and sentence representations but also a comprehension of their corresponding interaction. 

Existing methods on the VG task can be generally classified into three primary categories: (\uppercase\expandafter{\romannumeral1}) Proposal-based methods~\cite{Gao_Sun_Yang_Nevatia_2017, Xu_He_Plummer_Sigal_Sclaroff_Saenko_2018, Zhang_Peng_Fu_Luo_2019, Yuan_Ma_Wang_Liu_Zhu_2022}: Following the propose-and-rank pipeline, these methods first generate a set of proposals with predefined multi-scale anchors, then predict alignment scores for each proposal after aggregating contextual multi-modal information to identify the optimal matching moment, as depicted in Fig.~\ref{fig:teaser}(b). Notably, the performance of such proposal-based methods heavily relies on both the quantity and quality of predefined anchors, thereby imposing a substantial computational cost arising from the redundant proposal generation and matching processes. (\uppercase\expandafter{\romannumeral2}) Proposal-free methods~\cite{Zeng_Xu_Huang_Chen_Tan_Gan_2020, Nan_Qiao_Xiao_Liu_Leng_Zhang_Lu_2021, Zhang_Sun_Jing_Zhen_Zhou_Goh_2021}: As shown in Fig.~\ref{fig:teaser}(c), these methods directly regress the start and end timestamps from each frame or predict the probabilities for each frame to serve as the start and end frame, which is more efficient than the proposal-based methods. 

\begin{figure*}[t]
\centering
\includegraphics[width=0.91\linewidth]{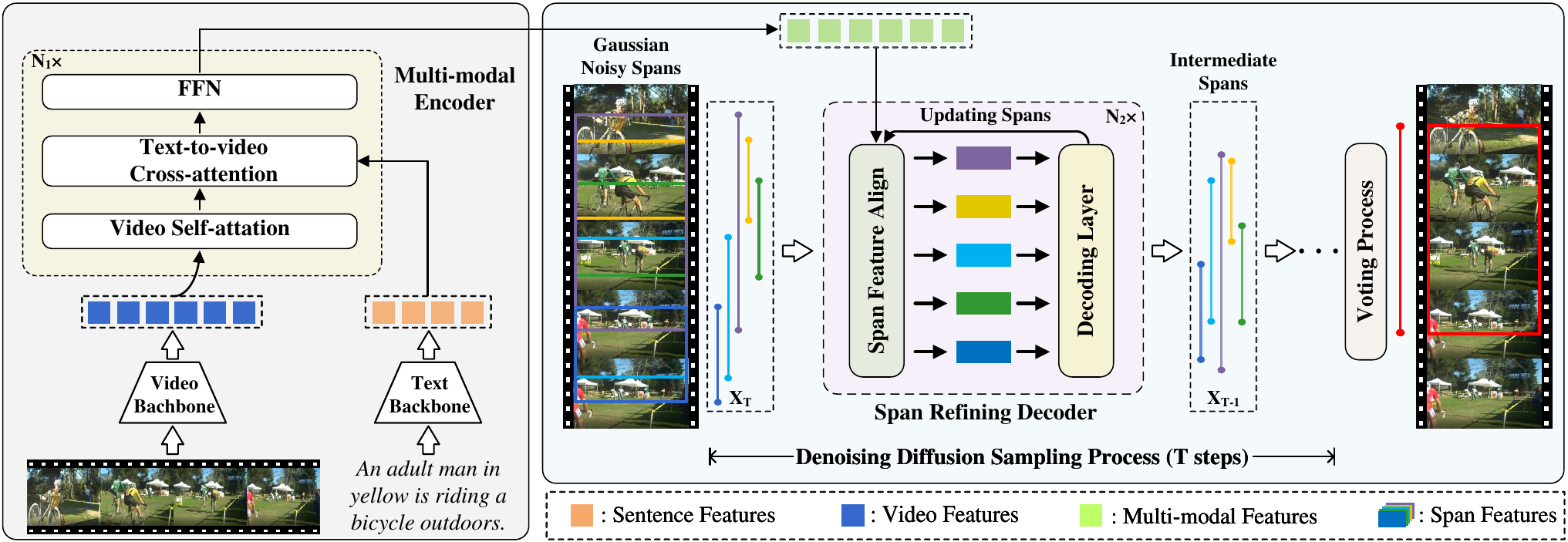}
\caption{Overview of DiffusionVG. Taking a video and a sentence query as input, DiffusionVG first extracts features from both modalities and facilitates interaction using a video-centered multi-modal encoder. Subsequently, the predictions of the target span are generated in a span refining decoder conditioned on the encoded multi-modal representations and iteratively refined through the reverse diffusion process. The final prediction of the target span is selected via a voting process.}
\label{fig:pipeline}
\end{figure*}

Nevertheless, incorporating frame-wise predictions poses challenges in optimization due to the extensive search space for moment prediction. Consequently, proposal-free methods tend to yield inferior performance.
(\uppercase\expandafter{\romannumeral3}) Transformer-based methods~\cite{Vaswani_Shazeer_Parmar_Uszkoreit_Jones_Gomez_Kaiser_Polosukhin_2017}: Some studies~\cite{Lei_Berg_Bansal_2021, liu2022umt, moon2023query} seek to employ transformer-based models on the VG task, combining both methodologies mentioned above by incorporating a set of learnable queries with regression heads to derive the target span. However, the fixed number of learnable queries tends to focus on specific video patterns and exploits the location bias inherent in the dataset~\cite{Gao_Sun_Yang_Nevatia_2017, Zhang_Peng_Fu_Luo_2019}, thereby limiting the model's generalizability. 
Inspired by the inherent characteristic of multi-step denoising process in diffusion models
~\cite{Sohl-Dickstein_Weiss_Maheswaranathan_Ganguli_2015, Ho_Jain_Abbeel_2020,chen2022diffusiondet}
, this paper presents DiffusionVG, a novel framework that employs diffusion models to formulate VG as a conditional generative task.
The central motivation behind DiffusionVG is to achieve the iterative refinement of the predicted target span by leveraging the advantage of denoising diffusion.
As illustrated in Fig.~\ref{fig:teaser}(d), the predicted target span in DiffusionVG is generated from Gaussian noise inputs conditioned on the video-sentence representations through the reverse diffusion process. 
To achieve this, the proposed DiffusionVG follows an encoder-decoder architecture~\cite{Vaswani_Shazeer_Parmar_Uszkoreit_Jones_Gomez_Kaiser_Polosukhin_2017}, wherein a video-centered multi-modal encoder facilitates the interaction between video and sentence features and a specialized span refining decoder is employed for generating the target span. 
During training, Gaussian noise with a predefined variance schedule is systematically added to the ground-truth span in the forward diffusion process, resulting in the production of noisy spans. 
Subsequently, the denoising decoder is learned to recover the original target span, conditioned on the video-sentence representations obtained from the multi-modal encoder.
At the inference stage, once the video-sentence representations are encoded, DiffusionVG generates the initial predicted spans from Gaussian noise inputs and iteratively refines the predictions through the learned reverse diffusion process. Finally, a voting strategy is employed to determine the best prediction among all generated spans.


In contrast to aforementioned methods, our proposed DiffusionVG eliminates the need for predefined anchors and concentrates all queries (noisy spans) on recovering a single target span, which significantly simplifies the optimization procedure. 
Moreover, once the DiffusionVG is trained, it has the flexibility to utilize any number of denoising sampling steps and queries in inference, irrespective of the configuration in the training stage.
This adaptability of DiffusionVG allows for optimizing the accuracy of target moment localization while maintaining a balance with inference speed.

To demonstrate the effectiveness of DiffusionVG, we conduct extensive experiments on the mainstream Charades-STA~\cite{Gao_Sun_Yang_Nevatia_2017}, ActivityNet Captions~\cite{Krishna_Hata_Ren_Li_Niebles}, and TACoS~\cite{Regneri_Rohrbach_Wetzel_Thater_Schiele_Pinkal_2013} datasets. Without bells and whistles, our DiffusionVG surpasses all existing well-designed state-of-the-art methods on all three benchmarks across all evaluation metrics.


Our \textbf{contributions} can be summarized as follows: (1) We innovatively formulate video grounding as a conditional generative task using diffusion models, enabling iterative refinement of predicted spans through the reverse denoising diffusion process. (2) We further design a video-centered encoder and a span refining decoder for our proposed DiffusionVG to optimize its compatibility with the VG task and enhance performance. (3) Extensive experiments on the Charades-STA, ActivityNet Captions, and TACoS benchmarks consistently demonstrate that DiffusionVG considerably outperforms existing state-of-the-art models across all evaluation metrics.





%
\section{Method}
\label{sec:method}


\subsection{Preliminary}
\label{sec:prelim}
\noindent \textbf{Video Grounding Task Formulation}. 
Given an untrimmed video $\boldsymbol{V}=\left\{v_m\right\}_{m=1}^M$ consisting of $M$ consecutive frames and a sentence query $\boldsymbol{Q}=\left\{q_n\right\}_{n=1}^N$ with $N$ words, a VG model aims to localize the start and end timestamps $\mathbf{z}_0 = \left(\tau_s, \tau_e\right)$ of the target moment (termed as target span) in $\boldsymbol{V}$ that best matches the query $\boldsymbol{Q}$, where $v_m$ demotes the $m$-th frame in $\boldsymbol{V}$ and $q_n$ demotes the $n$-th word in $\boldsymbol{Q}$. In our setting, VG is framed as a conditional generation task, where the target span is generated through the reverse process of diffusion models. To achieve this, a denoising network $\theta$ is adopted to recover the target span $\mathbf{z}_0$ from the noisy span $\mathbf{z}_t$, with the video $\boldsymbol{V}$ and query $\boldsymbol{Q}$ serving as conditions.

\noindent \textbf{Diffusion Model}.
Diffusion models~\cite{Sohl-Dickstein_Weiss_Maheswaranathan_Ganguli_2015, Ho_Jain_Abbeel_2020} are a class of likelihood-based generative models consisting of a forward process and a reverse process, both of which can be considered as a Markov chain. Formally, given a data sample $\mathbf{z}_0$ from distribution $q\left(\mathbf{z}_0\right)$ and a predefined noise schedule $\left\{\beta_1, \ldots, \beta_T\right\}$ for $T$ steps, the forward process transforms $\mathbf{z}_0$ to the noisy latent sample $\mathbf{z}_t$ for $t \in\{0,1,\ldots,T\}$ by gradually adding Gaussian noise to $\mathbf{z}_0$, formulated as:
\begin{equation}
\label{eq:forward}
q\left(\mathbf{z}_t \mid \mathbf{z}_0\right)=\mathcal{N}\left(\mathbf{z}_t ; \sqrt{\bar{\alpha}_t} \mathbf{z}_0,\left(1-\bar{\alpha}_t\right) \mathbf{I}\right),
\end{equation}
where $\bar{\alpha}_t =\prod_{k=1}^t \alpha_k=\prod_{k=1}^t\left(1-\beta_k\right)$ is commonly pre-calculated for efficient one-step sampling at an arbitrary time step $t$.
In the training stage, a neural network $\theta$ is learned to predict $\mathbf{z}_0$ from $\mathbf{z}_t$ as $\hat{\mathbf{z}}_0 = f_\theta\left(\mathbf{z}_t, t\right)$, which can be optimized with a $\ell_2$ loss function according to the derivation in~\cite{Ho_Jain_Abbeel_2020}: 
\begin{equation}
\mathcal{L}_{\text {obj }}=\frac{1}{2}\left\|f_\theta\left(\mathbf{z}_t, t\right)-\mathbf{z}_0\right\|^2.
\end{equation}
During inference, a data sample $\mathbf{z}_0$ can be derived from a Gaussian noise $\mathbf{z}_T$ with $\theta$ in an iterative manner. 
To speed up the sampling process, our method incorporates the DDIM updating rule presented in~\cite{Song_Meng_Ermon_2020}.

\begin{algorithm}[t]
\small
 \caption{Training}
 \label{alg:training}
 \Repeat {\normalfont{converged}}{
  Sample a video $\boldsymbol{V}$ and a query $\boldsymbol{Q}$ with target span $\mathbf{z}_0$\\
  Extract video features $\mathbf{\tilde{F}}^V$ and sentence features $\mathbf{\tilde{F}}^Q$ \\
  $\mathbf{H}$ = Encoding($\mathbf{\tilde{F}}^V, \mathbf{\tilde{F}}^Q$) \\
  \vspace{0.025cm}
  $t \sim$ Uniform $(\{1, \ldots, T\})$ \\
  $\epsilon \sim \mathcal{N}(\mathbf{0}, \mathbf{I})$ \\
  $\mathbf{z}_t =\sqrt{\bar{\alpha}_t} \mathbf{z}_0+\sqrt{1-\bar{\alpha}_t } \epsilon$ \\
  Recover $\mathbf{z}_0$ by computing $\hat{\mathbf{z}}_0 = f_{\theta}(\mathbf{z}_t, \mathbf{H}, t)$ \\
  Computing the span loss function for optimization
 }
\end{algorithm}

\subsection{DiffusionVG Model}
As shown in Fig.~\ref{fig:pipeline}, our DiffusionVG framework consists of two core components: multi-modal feature encoding and target span generation. In multi-modal feature encoding, we initially employ distinct uni-modal backbones to extract features from both video and sentence inputs. These uni-modal features are subsequently fused in the video-centered multi-modal encoder, resulting in text-enhanced video features. In target span generation, the text-enhanced video features serve as conditions to guide the denoising diffusion process for generating the target span. 

\noindent \textbf{Feature Extraction}.
For video feature extraction, we follow the previous work~\cite{Zhang_Peng_Fu_Luo_2019} and employ a pretrained 3D convolutional network 
to extract video features. Specifically, given an input video consisting of $M$ frames, we identically segment the video into small clips $\left\{v_i\right\}_{i=1}^{M/m}$, with each clip comprising $m$ frames. Next, we employ the pretrained 3D convolutional network to extract clip-level features. We perform a fixed-length sampling with an even stride $\frac{M}{m \cdot K}$ to obtain $K$ clip-level features. These fixed-length features are then fed into a fully connected (FC) layer for dimension reduction, resulting in the final video features $\mathbf{F}^V \in \mathbb{R}^{K \times d}$. For sentence feature extraction, we employ a standard pretrained language model 
to extract the word-level features from $\boldsymbol{Q}$, followed by dimension reduction via an FC layer to match the dimension of video features, resulting in $\mathbf{F}^Q \in \mathbb{R}^{N \times d}$. The specific feature encoders for each modality can be found in Appendix D.

\begin{algorithm}[t]
\small
 \caption{Inference}
 \label{alg:inference}
  Extract video features $\mathbf{\tilde{F}}^V$ and sentence features $\mathbf{\tilde{F}}^Q$ \\
  $\mathbf{H}$ = Encoding($\mathbf{\tilde{F}}^V, \mathbf{\tilde{F}}^Q$) \\
  \vspace{0.025cm}
 $\mathbf{z}_T \sim \mathcal{N}(\mathbf{0}, \mathbf{I}) \in \mathbb{R}^{N_q\times 2}$ \\
 $s$ is the reversed sampling steps of length $\tau$ with $s_\tau=T$\\
 \For{$i = \tau, \ldots, 1$}{
    \vspace{0.04cm}
    Compute $\hat{\mathbf{z}}_0$ with $f_{\theta}(\mathbf{z}_{s_i}, \mathbf{H}, s_i)$ \\
    $\mathbf{z}_{s_{i-1}} =$ ddim\_step ($\mathbf{z}_{s_{i}}, \hat{\mathbf{z}}_0, s_{i}, s_{i-1}$)
 }
 Select the final predicted span via voting\\
 \Return final span
\end{algorithm}


\begin{table*}[t!]
    \centering
    \setlength{\tabcolsep}{1.3mm}{
    \renewcommand\arraystretch {0.80}
    \scalebox{0.82}{
    \begin{tabular}{l|c|cccc|cccc|ccc}
    \toprule[1.5pt]
    \multirow{3}*{\textbf{Method}} & \multirow{3}*{Setting}  & \multicolumn{4}{c|}{\large{\textbf{Charades-STA}}} & \multicolumn{4}{c}{\large{\textbf{ActivityNet Captions}}} & \multicolumn{3}{c}{\large{\textbf{TACoS}}}  \\
    ~ & ~ & \multirow{2}*{\makecell{Video \\ Feature}} & R@1 & R@1 & R@1 & \multirow{2}*{\makecell{Video \\ Feature}} &  R@1 & R@1 & R@1 & \multirow{2}*{\makecell{Video \\ Feature}} &  R@1 & R@1 \\ 
    ~ & ~ & ~ &  IoU=0.3 & IoU=0.5 & IoU=0.7  & ~  & IoU=0.3 & IoU=0.5 & IoU=0.7  & ~  & IoU=0.3 & IoU=0.5 \\  
    \midrule[1.0pt]
    CTRL~\cite{Gao_Sun_Yang_Nevatia_2017} & PB & C3D  & - & 23.63 & 8.89  & C3D & - & 29.01 & 10.34 & C3D & 18.32 & 13.30  \\
    QSPN~\cite{Xu_He_Plummer_Sigal_Sclaroff_Saenko_2018} & PB & C3D & 54.70 & 35.60 & 15.80  & C3D & 45.30 & 33.26 & 13.43 &- & -&-  \\
    2DTAN~\cite{Zhang_Peng_Fu_Luo_2019} & PB & C3D & - & 39.81 & 23.25  & C3D & - & 44.51 & 27.38 & C3D& 37.29 & 25.32 \\
    SCDM~\cite{Yuan_Ma_Wang_Liu_Zhu_2022} & PB & I3D & - & 54.44 & 33.43  & C3D & 54.80 & 36.75 & 19.86 & -& -&- \\
    \midrule[1.0pt]
    DRN~\cite{Zeng_Xu_Huang_Chen_Tan_Gan_2020} & PF & I3D & - & 53.09 & 31.75 & C3D & - & 45.45 & 24.36 & C3D & & 23.17\\
    IVG-DCL~\cite{Nan_Qiao_Xiao_Liu_Leng_Zhang_Lu_2021} & PF & I3D & 67.63 & 50.24 & 32.88  & C3D & 63.22 & 43.84 & 27.10 & C3D & 38.79 & 28.89 \\
    VSLNet-L~\cite{Zhang_Sun_Jing_Zhen_Zhou_Goh_2021} & PF & I3D & 70.46 & 54.19 & 35.22 & I3D & 62.35 & 43.86 & 27.51 & I3D &47.66 & 36.34 \\
    \midrule[1.0pt]
    Moment-DETR*~\cite{Lei_Berg_Bansal_2021} & TF & VGG & 64.58 & 49.85 & 27.62 &I3D & 55.18 & 37.07 & 21.54 & C3D & 37.63 & 26.59\\
    UMT~\cite{liu2022umt} & TF & VGG & - & 48.44 & 29.76 &- & - & - & -  & - & - & - \\
    QD-DETR*~\cite{moon2023query}. & TF & VGG & 67.35 & 52.77 & 31.13 &I3D & 55.36 & 37.20 & 21.19  & C3D & 39.74 & 28.47\\
    \midrule[1.0pt]
    \textbf{DiffusionVG} (1-step) & GN & I3D & 73.17 & 61.24 & 38.97 & C3D & 63.95 & 43.87 & 25.45 & C3D & 45.12 & 37.05\\
    \textbf{DiffusionVG} (5-step) & GN & \makecell{I3D \\ VGG}  & \makecell{\textbf{76.53} \\ \underline{70.38}} & \makecell{\textbf{62.30} \\ \underline{57.13}} & \makecell{\textbf{40.05} \\ \underline{35.06}} & C3D & \textbf{65.02} & \textbf{47.27} & \textbf{27.87} & C3D & \textbf{47.93} & \textbf{38.02} \\
    
    \bottomrule[1.5pt]
    \end{tabular}}}
    \caption{Performance comparison between DiffusionVG and state-of-the-art methods on Charades-STA, ActivityNet Captions, and TACoS datasets. \textbf{PB} denotes the proposal-based setting, \textbf{PF} denotes the proposal-free setting, \textbf{TF} denotes the Transformer-based setting and \textbf{GN} denotes our unique generative setting. (Best results are in \textbf{bold} and the \underline{underline} denotes the best performance with VGG features, * refers to reproduced results from official implementation.)}
    \label{tab:compare}
\end{table*}

\noindent \textbf{Video-centered multi-modal Encoding}.
To enhance the representation of the target moment in the video, our DiffusionVG employs a video-centered multi-modal encoder that consistently integrates sentence information into the contextualized video features. Specifically, our video-centered multi-modal encoder comprises a stack of $N_1$ post-normalization Transformer encoder layers with modification. Each encoder layer consists of a video self-attention layer, a text-to-video cross attention layer and a feed-forward layer. With the video and sentence features available, we first add a sine positional embedding $\mathbf{P}^V$ to video features and a learnable position embedding $\mathbf{P}^Q$ to sentence features~\cite{Vaswani_Shazeer_Parmar_Uszkoreit_Jones_Gomez_Kaiser_Polosukhin_2017}
, resulting in $\mathbf{\tilde{F}}^V$ and $\mathbf{\tilde{F}}^Q$, which then serve as inputs to the video-centered multi-modal encoder. Within the video self-attention layer, intra-modal self-attention is performed between clip-level features to derive the contextualized video features. 
In the text-to-video cross-attention layer, the query for cross-attention is obtained from video features, while key-value pairs are sourced from sentence features. Therefore, only the video features are updated in this layer. Furthermore, a feed-forward network (FFN) is exclusively applied to the video features. The output of the last layer of FFN in our video-centered multi-modal encoder yields the text-enhanced video features, denoted as $\mathbf{H} \in \mathbb{R}^{K \times d}$, which serve as conditions to guide the generation of the target span in the subsequent denoising diffusion process.

\noindent \textbf{Span Refining in Decoding}.
During the target span generation process, DiffusionVG takes $N_q$ Gaussian noise spans as inputs and iteratively refines them through the reverse diffusion process, with conditioning on the text-enhanced video features $\mathbf{H}$. 
In order to enhance the refinement within the generation process, we incorporate a span refining decoder that facilitates further refinement of span predictions at each decoding layer. Specifically, each layer of our span refining decoder is composed of a cross-attention and a feed-forward sub-layer, which takes the noisy spans, time step $t$ and the encoded text-enhanced video-centered features $\mathbf{H}$ as inputs. We crop the segments of $\mathbf{H}$ corresponding to the noisy spans $\mathbf{z}_t$ and apply a soft-pooling operation to aggregate the features in each segment as span proposals for decoding. 
The output of every decoder layer estimates the differences between the input noisy spans and the target span, which are subsequently utilized to update the noisy spans that serve as the inputs of the next decoder layer. 
Through the updates in the stack of $N_2$ decoder layers, the output spans from the last decoder layer corresponds to the predicted target spans at time step $t$. More information and detailed calculations for our span refining decoder can be found in Appendix A.

\subsection{Training Strategy}
During training, our DiffusionVG produces the noisy spans by adding Gaussian noise to the ground-truth span and then learns the reverse process conditioned on the video-sentence representations, as depicted in Algorithm.~\ref{alg:training}. 

\noindent \textbf{Noisy Spans}.
We introduce Gaussian noises to the ground-truth span to obtain the noisy spans. The schedule of the noises is regulated by the predefined hyper-parameter $\alpha_t$. In line with~\cite{chen2022diffusiondet}, we employ a monotonically decreasing cosine schedule at different time step $t$. As a data augmentation approach, we replicate the ground-truth span $N_q$ times with various time step $t$ and add independent Gaussian noises to them. Through the utilization of parallel computing, this approach accelerates the convergence of training process. Following~\cite{chen2022diffusiondet}, we enhance the signal-to-noise ratio (SNR) in the forward diffusion process by multiplying a scaling factor $\lambda$ to the ground-truth span before injecting noises. This technique has demonstrated promising potential in improving the performance of DiffusionVG when an appropriate $\lambda$ is utilized. 

\noindent \textbf{Loss Function}.
The output of DiffusionVG is solely comprised of the predicted target spans, each of which is supervised by the same ground-truth span. We apply a combination of a $\ell_1$ loss and a generalized IoU (GIoU)~\cite{Rezatofighi_Tsoi_Gwak_Sadeghian_Reid_Savarese_2019} loss for optimization, formulated as:
\begin{equation}
\label{eq:loss}
\mathcal{L}=\lambda_{\ell 1} \cdot \mathcal{L}_{\ell 1}(\hat{\mathbf{z}}_0, \mathbf{z}_0)+\lambda_{\text {giou }} \cdot \mathcal{L}_{\text {giou }}(\hat{\mathbf{z}}_0, \mathbf{z}_0),
\end{equation}
where $\lambda_{\ell 1}$ and $\lambda_{giou}$ are the balancing weights of each component. Moreover, auxiliary loss~\cite{Carion_Massa_Synnaeve_Usunier_Kirillov_Zagoruyko_2020} is incorporated in every decoder layer to accelerate convergence.

\subsection{Inference}
In inference, our proposed DiffusionVG takes Gaussian noises as inputs and generates the target spans with iterative refinement through the reverse diffusion process, as depicted in Algorithm.~\ref{alg:inference}. The whole generation process is conditioned on the encoded video-sentence representations. It is worth noting that once DiffusionVG finishes training, it becomes flexible to utilize an arbitrary number of sampling steps and queries (noisy spans inputs) in inference, regardless of the configurations in the training stage. 

\noindent \textbf{Sampling Process}.
In each sampling step $t$, the noisy spans obtained from the previous step (or Gaussian noise inputs) are fed into the span refining decoder to generate the predictions of the target span. Once the predicted target spans at current step $t$ are acquired, we employ the DDIM~\cite{Song_Meng_Ermon_2020} updating strategy to produce the refined noisy spans for the next step. The final predicted target spans of DiffusionVG are obtained from the last denoising sampling step.

\noindent \textbf{Voting Strategy}.
Unlike the settings in object detection, our decoder does not estimate the scores of each generated span in the reverse diffusion process, since all queries aim to reconstruct the same target span rather than the set matching problem in~\cite{Lei_Berg_Bansal_2021, chen2022diffusiondet, Carion_Massa_Synnaeve_Usunier_Kirillov_Zagoruyko_2020}. In order to determine the best prediction among all generated spans, we present a voting strategy where all spans participate in a voting process with each other to determine the scores for each candidate. Specifically, we calculate the IoU of each generated span with all other spans and use a sum of these IoUs as its score. The span with the highest score is selected as the final prediction. The validation of the voting strategy can be found in Appendix C.



\section{Experiments}
\label{sec:exp}

In this section, we provide a comprehensive evaluation of our proposed DiffusionVG and compare it with state-of-the-art methods. We start by introducing the datasets and evaluation metrics. Next, we compare DiffusionVG with existing methods in VG and conduct extensive ablation studies. Finally, we showcase the qualitative results obtained from DiffusionVG. The implementation details can be found in Appendix D.

\begin{figure}[t]
\centering
\includegraphics[width=7.3cm]{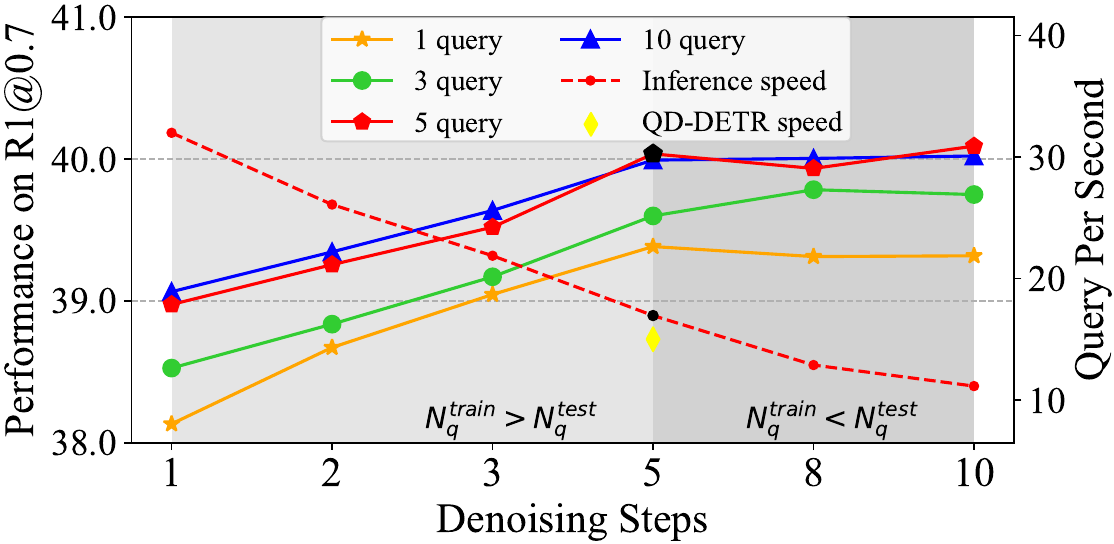}
\caption{Evaluating the impact of \textbf{sampling steps} and \textbf{number of queries} on model performance and inference speed (5 queries). All experiments are conducted on Charades-STA test set. Black markers denote the default setting.}
\label{fig:sampling_steps}
\end{figure}



\subsection{Datasets and Metrics}
\label{sec:datasets}
\noindent \textbf{Datasets}. To demonstrate the effectiveness of our method, we conduct experiments on three mainstream benchmarks: Charades-STA~\cite{Gao_Sun_Yang_Nevatia_2017}, ActivityNet Captions~\cite{Krishna_Hata_Ren_Li_Niebles} and TACoS~\cite{Regneri_Rohrbach_Wetzel_Thater_Schiele_Pinkal_2013}. 
\textbf{Charades-STA} comprises a total of 6,672 videos depicting daily indoor activities, which is annotated with sentence descriptions focusing on specific temporal moments. This dataset includes 12,408 (from 5,338 videos) moment-sentence pairs for training and 3,720 (from 1,334 videos) moment-sentence pairs for testing. \textbf{ActivityNet Captions} consists of around 20k videos in diverse human activities. 
Following the settings in previous works~\cite{Zhang_Peng_Fu_Luo_2019, Zhang_Sun_Jing_Zhen_Zhou_Goh_2021}, we use its original training set for training, val\_1 for validation and val\_2 for testing, which have 37,417, 17,505 and 17,031 moment-sentence pairs. \textbf{TACoS} contains 127 videos  that center on indoor cooking scenes, partitioned into 10,146, 4,589, and 4,083 moment-sentence pairs annotated for training, validation, and testing, respectively.

\noindent \textbf{Evaluation Metrics}. 
We follow~\cite{Gao_Sun_Yang_Nevatia_2017} and evaluate DiffusionVG with the commonly used ``Rank@$n$, IoU=$m$" (R$n$@$m$) metric. The R$n$@$m$ metric calculates the percentage of the queries that have at least one predicted moment whose temporal Intersection over Union (IoU) is larger than the threshold $m$ with the ground-truth moment among the top-$n$ predicted moments. In our experiments, we employ the most commonly adopted R$1$@$0.3$, R$1$@$0.5$, and R$1$@$0.7$ metrics for both Charades-STA and ActivityNet Captions datasets, while only adopting the first two metrics for TACoS since this is consistent with most existing methods~\cite{Gao_Sun_Yang_Nevatia_2017, Zhang_Peng_Fu_Luo_2019,Zeng_Xu_Huang_Chen_Tan_Gan_2020, Nan_Qiao_Xiao_Liu_Leng_Zhang_Lu_2021}.

\begin{table}[t]
    \centering
    \footnotesize
    \renewcommand\arraystretch {0.8}
    \begin{tabular}{cc|ccc}
      \toprule[1pt]
      VcE & SRD & R1@0.3 & R1@0.5 & R1@0.7 \\ 
      \midrule[1pt]
         ~ & ~  & 73.92 & 60.27 & 35.78 \\ 
         \CheckmarkBold  & ~ & 75.93 & 61.48 & 37.32 \\ 
         ~ & \CheckmarkBold & 75.86 & 61.74 & 38.27 \\ 
         \CheckmarkBold & \CheckmarkBold & \textbf{76.53} & \textbf{62.30} & \textbf{40.05} \\ 
      \bottomrule[1pt]
  \end{tabular}
  \caption{Ablation study on our \textbf{video-centered encoder} (VcE) and \textbf{span refining decoder} (SRD).}
  \label{tab:ablation}
\end{table}

\subsection{Comparison with State-of-the-arts}
\label{sec:comparison}
To validate the effectiveness of our proposed DiffusionVG, we compare it with state-of-the-art VG methods on mainstream datasets, as shown in Table~\ref{tab:compare}. The compared methods can be divided into three groups: (1) \textbf{Proposal-based Methods}: CTRL~\cite{Gao_Sun_Yang_Nevatia_2017}, QSPN~\cite{Xu_He_Plummer_Sigal_Sclaroff_Saenko_2018}, 2DTAN~\cite{Zhang_Peng_Fu_Luo_2019} and SCDM~\cite{Yuan_Ma_Wang_Liu_Zhu_2022}. 
These methods employ a propose-and-rank pipeline, wherein a set of moment proposals are pre-sampled from the input video and matched with the sentence query to identify the optimal target moment.
(2) \textbf{Proposal-free Methods}: DRN~\cite{Zeng_Xu_Huang_Chen_Tan_Gan_2020}, IVG-DCL~\cite{Nan_Qiao_Xiao_Liu_Leng_Zhang_Lu_2021} and VSLNet-L~\cite{Zhang_Sun_Jing_Zhen_Zhou_Goh_2021}. These methods perform direct regression of the start and end timestamps from each frame or predict the probabilities for each frame to be the start or end frame.
(3) \textbf{Transformer-based Methods}: Moment-DETR~\cite{Lei_Berg_Bansal_2021}, UMT~\cite{liu2022umt} and QD-DETR~\cite{moon2023query}. These methods employ Transformer-based architecture on the VG task. To further demonstrate the effectiveness of the iterative refinement in our DiffusionVG, we compare it with DETR-like baselines under the same parameter settings in Appendix B.

\noindent \textbf{Main Results}. 
As illustrated in Table~\ref{tab:compare}, DiffusionVG outperforms existing methods across all evaluation metrics on all three benchmarks, establishing new state-of-the-art performances for the VG task. Notably, DiffusionVG demonstrates an average absolute improvement of 3.09\% compared to the previously best-performing approaches. The absolute gain on the R$1$@$0.5$ metric for the Charades-STA dataset is particularly remarkable, with a significant improvement of 7.86\%.  It is worth mentioning that our proposed DiffusionVG does not rely on intricately-designed model architectures or specialized learning strategies, yet it surpasses all its discriminative learning-based counterparts such as proposal-based 2D-TAN and proposal-free VSLNet-L. This highlights the potential of diffusion-based generative models in tackling the VG task, as the iterative framework allows for continuous refinement of the target moment prediction. 




\subsection{Ablation Study}
\label{sec:ablation}
In this section, we conduct comprehensive ablation experiments on the Charades-STA dataset to study DiffusionVG in detail. Specifically, we evaluate the impact of different sampling steps, and investigate the effectiveness of our proposed video-centered encoder and span-refining decoder. More ablation studies are provided in Appendix C.


\noindent \textbf{Sampling Steps}.
As shown in Fig.~\ref{fig:sampling_steps}, increasing the number of sampling steps leads to a gradual improvement in model performance. However, this improvement comes at the cost of slower inference speed.
In the case of utilizing 5 queries as an example, when we employ only one step to generate the target span from Gaussian noise inputs, the model witnesses a substantial decline in performance (1.06\% and 1.08\% in R1@0.5 and R1@0.7 respectively). This observation demonstrates the effectiveness of the iterative refining process in DiffusionVG.
Note that under the default setting of 5 sampling steps, our DiffusionVG demonstrates a faster inference speed compared to the state-of-the-art Transformer-based method~\cite{moon2023query}.


\noindent \textbf{Encoder and Decoder Design}. 
To demonstrate the effectiveness of our proposed video-centered encoder and span refining decoder in DiffusionVG, we compare them with a standard multi-modal Transformer encoder-decoder architecture. In the standard Transformer, the encoder facilitates the interaction between multi-modal information, while the decoder ensures continuous attention of queries to the encoded multi-modal features. As shown in Table~\ref{tab:ablation}, both our video-centered encoder and span refining decoder contribute to the improved performance of DiffusionVG when compared to the standard counterparts. Furthermore, the improvements brought by both modules are compatible with each other, leading to further enhancement of the overall performance. 

\begin{figure}[t]
\begin{center}
\includegraphics[width=0.44\textwidth]{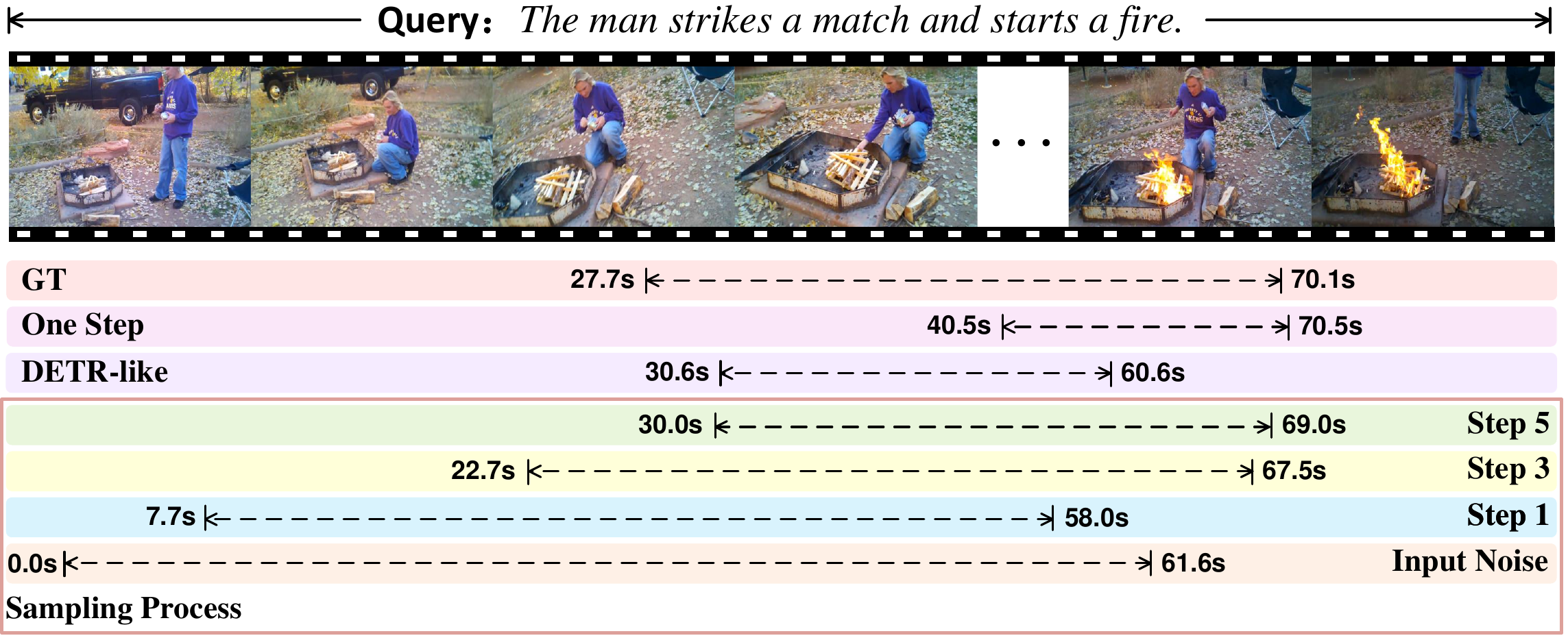}
\vspace{-0.3cm}
\caption{A visualized example of the proposed DiffusionVG on ActivityNet Captions dataset (only one query is adopted).}
\label{fig:vis}
\end{center}
\end{figure}

\subsection{Visualization} 
\label{sec:visualization}
To qualitatively highlight the effectiveness of the iterative refining process in the multi-step generation approach, we visualize a prediction comparison between the one-step and multi-step sampling strategies in DiffusionVG on the ActivityNet Captions dataset, as shown in Fig.~\ref{fig:vis}. 
We can observe that the one-step DiffusionVG model can understand the semantics of both video and sentence contents and localize a video segment that, to some extent, relates to the sentence query. However, its predicted moment significantly deviates from the ground truth. In contrast, utilizing a multi-step generation strategy enables DiffusionVG to perform iterative refinement on the predictions by mitigating the disparities between noisy spans and the target span, thereby yielding an improved grounding result compared to the one-step approach. More visualizations can be found in Appendix E.

\label{sec:implementation}

\section{Related Work}
\label{sec:related}

\subsection{Video Grounding (VG)}
As introduced by~\cite{Gao_Sun_Yang_Nevatia_2017,Hendricks_Wang_Shechtman_Sivic_Darrell_Russell_2017}, this task aims to retrieve a target moment from an untrimmed video that semantically corresponds to a given sentence query. To achieve precise localization of the desired moment via language descriptions, it is essential for a VG model to establish the semantic alignment between the video and sentence inputs. Early methods mainly resort to the propose-and-rank pipeline~\cite{Gao_Sun_Yang_Nevatia_2017, Xu_He_Plummer_Sigal_Sclaroff_Saenko_2018, Zhang_Peng_Fu_Luo_2019, Yuan_Ma_Wang_Liu_Zhu_2022}, which produce a set of moment proposals with predefined multi-scale anchors in the initial stage. After aggregating contextual multi-modal information, the model can predict alignment scores for each proposal and find the best matching moment. While these proposal-based methods provide reliable results, they have to perform cumbersome proposal generation and pair all proposals with the given query, which contribute to heavy computation cost. In contrast, recently proposal-free methods~\cite{Zeng_Xu_Huang_Chen_Tan_Gan_2020, Zhao_Zhao_Zhang_Lin_2021, Zhang_Sun_Jing_Zhen_Zhou_Goh_2021} have been developed to directly regress the start and end timestamps or estimate the probabilities for each frame to be the start and end frame, which can efficiently avoid the time-consuming proposal generating and ranking processes. Despite the efficiency advantages of the proposal-free methods, the integration of frame-wise predictions presents optimization challenges from the extensive search space, making these methods result in inferior performance. 
Recently, several studies~\cite{Lei_Berg_Bansal_2021, liu2022umt, moon2023query} attempt to leverage transformer-based models for the VG task, which incorporates a set of learnable queries with regression heads to determine the target span. However, the drawback of using a fixed number of learnable queries is its tendency to focus on specific video patterns and exploit the inherent location bias within the dataset~\cite{Gao_Sun_Yang_Nevatia_2017, Zhang_Peng_Fu_Luo_2019}. Consequently, this approach constrains the model's ability to generalize across diverse scenarios.
In contrast to existing methods, this paper presents DiffusionVG, an innovative framework that pioneers the exploration of diffusion-based generative models in VG.

\subsection{Diffusion Model}
Originally introduced in~\cite{Sohl-Dickstein_Weiss_Maheswaranathan_Ganguli_2015} and improved by~\cite{Song_Ermon_2019, Ho_Jain_Abbeel_2020, Song_Sohl-Dickstein_Kingma_Kumar_Ermon_Poole_2021}, diffusion models consist of a forward diffusion process (add noises to data) and a reverse denoising process (recover data from noises). Due to the robust generation and generalization capabilities, diffusion models have exhibited remarkable achievements in the fields of both computer vision~\cite{Fan_Chen_Chen_Cheng_Yuan_Wang_2022,Saharia_Chan_Saxena_Li_Whang_Denton_Kamyar_Ghasemipour_Karagol_Mahdavi_et_al,Harvey_Naderiparizi_Masrani_Weilbach_Wood_2022} and natural language processing~\cite{Austin_Johnson_Ho_Tarlow_Berg_2021, Gong_Li_Feng_Wu_Kong_2022, Li_Thickstun_Gulrajani_Liang_Hashimoto_2022}. 

The majority of studies on diffusion models focus on their generative aspects, overlooking their discriminative capability on perception tasks. Some pioneer studies~\cite{baranchuk2021label, wolleb2022diffusion, amit2021segdiff} have attempted to employ diffusion models for image segmentation tasks.
\cite{chen2022diffusiondet} manages to apply diffusion models for object detection, where the bounding boxes of objects are generated from Gaussian noise inputs conditioned on the image features. In this paper, we take the first step in employing diffusion models in video grounding by formulating it as a conditional generative task, where the target span is predicted in a generative manner conditioned on video-sentence multi-modal representations.

\subsection{Video-Language Transformers}
Simultaneously understanding representations from both video and language is the major challenge for video-language models. Inspired by the great success of Transformers~\cite{Vaswani_Shazeer_Parmar_Uszkoreit_Jones_Gomez_Kaiser_Polosukhin_2017} in natural language processing, many studies~\cite{Sun_Myers_Vondrick_Murphy_Schmid_2019, Li_Chen_Cheng_Gan_Yu_Liu_2020,lei2021less, xu2021vlm} extend the Transformer-based architecture to video-language tasks, including video captioning~\cite{Zhou_Zhou_Corso_Socher_Xiong_2018}, video question answering~\cite{Lei_Yu_Bansal_Berg_2018} and text-to-video retrieval~\cite{Jun_Tao_Ting_Yong_2016}. Our proposed DiffusionVG also utilizes the architecture of Transformers and incorporates specialized encoder and decoder to generate the target span through the denoising diffusion process.

\section{Conclusion}
\vspace{-0.15cm}
\label{sec:conclusion}


In this paper, we propose DiffusionVG, a pioneering approach that addresses the video grounding task from the perspective of conditional generation. By employing a multi-step generation process through the denoising diffusion process in DiffusionVG, the predicted spans undergo iterative refinement starting from Gaussian noise inputs. Extensive experiments on mainstream benchmarks demonstrate the superiority of DiffusionVG over existing state-of-the-art methods.
\vspace{-0.25cm}




\clearpage
\bibliographystyle{IEEEbib}
\bibliography{icme2023template}
\clearpage
\appendix
\section*{Appendix}
\label{sec:appendix}

\subsection*{A. Decoder in Details}
\begin{figure}[h]
\centering
\includegraphics[width=1\linewidth]{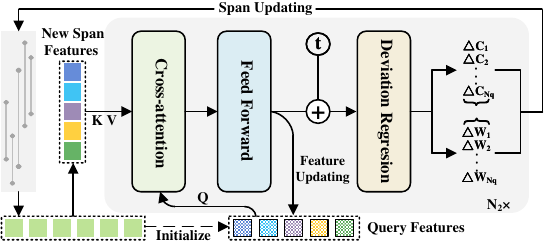}
\caption{Illustration of the span refining decoder. (Span location embedding is not illustrated.)}
\label{fig:decoder}
\end{figure}

\noindent In this section, we provide a detailed introduction to our proposed span refining decoder in DiffusionVG. As illustrated in Fig.~\ref{fig:decoder}, our span refining decoder takes $N_q$ noisy spans (or Gaussian noises) as inputs and generates the predicted target spans at time step $t$, with conditioning on the text-enhanced video features $\mathbf{H}$.

Each layer in our span refining decoder consists of a cross-attention and an FFN sub-layers. We begin with obtaining the span features in decoding stages (similar to the learnable queries in DETR~\cite{Carion_Massa_Synnaeve_Usunier_Kirillov_Zagoruyko_2020}), in other words, how to align a noisy span with the encoded multi-modal features. First of all, we crop the feature segment of $\mathbf{H}$ that corresponds to the $i$-th noisy span $\mathbf{z}_t^i$. Considering the varying lengths of feature segments within each input noisy span, we apply a soft-pooling operation to aggregate the features in each segment. Specifically, supposing the length of a feature segment for a noisy span $\mathbf{z}_t^i$ is $n_t^i$, we compute the weights of these $n_t^i$ clip-level features in the segment using a fully connected (FC) layer $\mathbf{W}_F$ followed by a softmax operation. By performing a weighted sum of these clip-level features, we obtain the initial span features of each input noisy span. To incorporate location information into the initial span features, we apply an FC layer $\mathbf{W}_E$ to project $\mathbf{z}_t^i$ onto the dimension of corresponding features as the location embedding (Note that we do not visualize the location embedding in Fig.~\ref{fig:decoder}). This location embedding is then added to the initial span features, providing the final span features $\mathbf{\bar{F}}_{\mathbf{z}_t^i}$ for the input noisy span $\mathbf{z}_t^i$. The entire process can be formulated as follows: 
\begin{align}
 &\text{Emd}_{\mathbf{z}_t^i} = \mathbf{W}_E(\mathbf{z}_t^i),  \\
 &\mathbf{F}_{\mathbf{z}_t^i} = \text{crop}(\mathbf{H},\mathbf{z}_t^i),  \\
 &\mathbf{w}_t^i = \text{softmax}(\mathbf{W}_F(\mathbf{F}_{\mathbf{z}_t^i})), \\[-2mm]
 &\mathbf{\bar{F}}_{\mathbf{z}_t^i} = \sum_{k=1}^{n_t^i} \mathbf{F}_{\mathbf{z}_t^i}[k] \cdot \mathbf{w}_t^i[k] + \text{Emd}_{\mathbf{z}_t^i} .
\end{align}
The span features obtained from the input noisy spans $\mathbf{z}_t$ (from last DDIM step or Gaussian noise inputs) are designated as \textbf{query features}. In every decoder layer, we update the query features through a cross-attention with the new span features obtained from new noisy spans updated in the previous decoder layer, as illustrated in the cross-attention operation in Fig.~\ref{fig:decoder}. The process of acquiring the new span features from the new noisy spans is analogous to the derivation of the query features from the input noisy spans. Notably, in the first decoder layer where the updated noisy spans are not available, we replace the cross-attention with a self-attention layer for updating the initial query features. Subsequently, the query features are further updated using an additive feed-forward layer, similar to the standard Transformer decoder layer. The time step $t$ is projected into a time embedding through a sine positional embedding~\cite{Vaswani_Shazeer_Parmar_Uszkoreit_Jones_Gomez_Kaiser_Polosukhin_2017} followed with a linear layer and added to the query features. In order to assess the deviation between current noisy spans and the target span, a linear layer is employed to estimate their differences in both centers and widths, which are subsequently utilized to update the noisy spans. Finally, the updated spans from the last decoder layer serve as predictions for the target span at time step $t$. During training, we directly optimize DiffusionVG by computing the loss function in Eq.3 (Method section) with the predicted target spans. In inference, we employ the DDIM updating rule to derive $\mathbf{z}_{t-\tau}$ that serves as the inputs of the next denoising step, where $\tau$ is the sampling stride.

\subsection*{B. Comparison with DETR-like Baselines}
To demonstrate the superiority of DiffusionVG over pure Transformer-based baselines, we compare it with DETR-like~\cite{Carion_Massa_Synnaeve_Usunier_Kirillov_Zagoruyko_2020} multi-modal Transformer models under the same parameter settings, except for the learnable queries as the decoder inputs for DETR-like models.

As indicated in Table.~\ref{tab:detr}, our DiffusionVG outperforms the DETR-like baselines across all evaluation metrics in the VG task, utilizing the same parameter settings. 
It is worth noting that even without employing multiple sampling steps, our DiffusionVG significantly outperforms DETR-like baselines by a large margin (9.95\% in R1@0.5 and 8.79\% in R1@0.7) while employing the same number of 5 queries.
We posit that this may due to the dilemma of defining the optimal learning objective for all learnable queries in DETR-like models. 
Specifically, when employing just one query, it may conform to a particular pattern in aligning video and sentence features, resulting in limited sensitivity to the rich representations of both videos and sentences. 
However, utilizing multiple queries presents the challenge of determining an appropriate learning objective for all predictions, since the rest predictions apart from the best-matched one may be considered as ``sub-optimal'' answers without being strictly incorrect. 
In contrast, the learning objective for all predictions in DiffusionVG is the recovery of a single ground-truth span, thus allowing flexibility in the number of queries. 
The utilization of multiple queries during the training stage can be considered a form of data augmentation. 
Thanks to the iterative refinement through the reverse diffusion process and the voting strategy, our DiffusionVG can precisely obtain the target span candidates and accurately select the best one.

\begin{table}[t]
    \vspace{0.18cm}
    \centering
    \large
    \resizebox{0.9\linewidth}{!}{
    \begin{tabular}{ccc|ccc}
      \toprule[1.5pt]
      Architecture & $N_s$ & $N_q$ & R1@0.3 & R1@0.5 & R1@0.7 \\ 
      \midrule[1pt]
        DETR-like & - & 1 & 66.18 & 51.94 & 29.01  \\ 
        DETR-like & - & 5 & 66.88 & 52.45 & 31.26 \\ 
        \midrule[1pt]
        DiffusionVG & 1 & 1 & 74.22 & 60.18 & 38.13 \\ 
        DiffusionVG & 1 & 5 & 73.17 & 61.24 & 38.97  \\ 
        DiffusionVG & 5 & 1 & 74.28 & 61.48 & 38.64 \\ 
        DiffusionVG & 5 & 5 & \textbf{76.53} & \textbf{62.30} & \textbf{40.05} \\ 
      \bottomrule[1.5pt]
  \end{tabular}}
  \vspace{-0.2cm}
  \caption{\textbf{Comparison with DETR-like baselines}. $N_s$ denotes the number of sampling steps for DiffusionVG and $N_q$ denotes the number of queries utilized in both models. Experiments are conducted on the Charades-STA dataset. }
  \label{tab:detr}
\vspace{-0.4cm}
\end{table}

\subsection*{C. More Ablation Study}
In this section, we provide additional ablation studies to further analyze each component in our DiffusionVG. This includes the impact of \textbf{number of queries} on task performance, the the effects of \textbf{signal scaling} factor $\lambda$, the impact of sampling steps and number of queries on \textbf{inference speed}, the influence of various \textbf{span representations} in decoder, the \textbf{stability} and \textbf{compatibility} of our DiffusionVG and the evaluation of the \textbf{voting strategy}. 

\vspace{0.1cm}
\noindent \textbf{1. Number of Queries}. 
Intuitively, a single query (input noisy span) suffices for recovering the target span in DiffusionVG. However, as shown in Fig.3 (Experiments section), the model performance can be enhanced as the number of queries increases in a certain range (from 1 to 5). We hypothesize that utilizing a limited number of queries contributes to the rising uncertainty through the generation process. Moreover, beyond a certain threshold, increasing the number of queries can have adverse effects on model performance (from 5 to 10). This may be attributed to the concurrent rise in low-quality predictions, leading to disturbances in the voting process for determining the best prediction.

\vspace{0.1cm}
\noindent \textbf{2. Signal Scaling}.
In Table.~\ref{tab:signal}, we study the impact of different signal scaling factors that control the signal-to-noise ratios (SNR) of the forward diffusion process. Unexpectedly, our model exhibits apparent sensitivity to the SNR, resulting in significant performance fluctuations with varying scaling factors. Nevertheless, we argue that the choice of the best scaling factor heavily relies on the specific task. For instance, most image generation tasks~\cite{Ho_Jain_Abbeel_2020, Dhariwal_Nichol_2021} prefer the value of 1.0, while it is 0.1 for panoptic segmentation~\cite{Chen_Li_Saxena_Hinton_Fleet_2022} and 2.0 for object detection~\cite{chen2022diffusiondet}. In line with the latter, our DiffusionVG showcases the best performance with the scaling factor of 2.0.

\begin{table}[h]
    \centering
    \resizebox{.85\linewidth}{!}{
    \scriptsize
    \renewcommand\arraystretch{0.85}
    \begin{tabular}{c|ccc}
      \toprule[1pt]
      Scaling & R1@0.3 & R1@0.5 & R1@0.7 \\ 
      \midrule[1pt]
        0.5  & 75.53 & 61.24 & 37.85 \\ 
        1.0  & 74.13 & 59.93 & 35.23 \\ 
        2.0  & \textbf{76.53} & \textbf{62.30} & \textbf{40.05} \\ 
        3.0  & 73.81 & 60.25 & 35.79 \\ 
      \bottomrule[1pt]
  \end{tabular}}
  \caption{\textbf{Signal scaling factor}. Impact of different signal-to-noise ratios on model performance.}
  \label{tab:signal}
\end{table}

\begin{figure}[h]
\centering
\includegraphics[width=8cm]{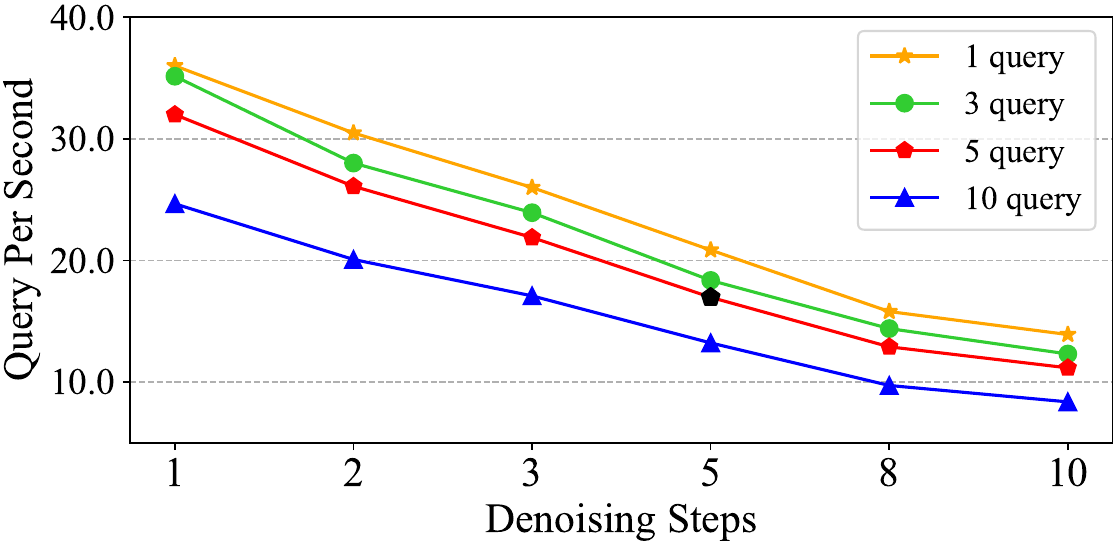}
\caption{Impact of \textbf{sampling steps} and \textbf{number of queries} on inference speed. All experiments are conducted on the test set of Charades-STA. (Default in black marker)}
\label{fig:sampling_speeds}
\end{figure}

\noindent \textbf{3. Balancing performance and speed}. 
As depicted in Fig.\ref{fig:sampling_speeds}, utilizing more sampling steps and number of queries both result in a compromise on the inference speed. Additionally, increasing the number of queries from 1 to 5 only slightly compromises the inference speed compared to the effect of increasing sampling steps. This discrepancy stems from the sequential execution inherent in the diffusion sampling process, while the utilization of multiple queries capitalizes on parallel computing, resulting in enhanced efficiency. Considering the trade-off between achieving better performance while maintaining efficiency, we set both the sampling steps and number of queries to 5 as default values in all experiments.

\noindent \textbf{4. Span Representations}.
In this section, we explore multiple approaches to represent the noisy spans (i.e., the way to align the noisy spans with the encoded multi-modal features $\mathbf{H}$) in our span refining decoder, including: (\textbf{1. feature}) We directly utilize the weighted sum of the features in the segment corresponding to the noisy span; (\textbf{2. cat-fn}) We first concatenate the weighted sum of segment features with the span location embedding, then apply an FC layer to project the concatenated features to the dimension of $\mathbf{H}$ for decoding; (\textbf{3. add}) The default setting in DiffusionVG, which adds the span location embedding to the weighted sum of segment features. As shown in Table.~\ref{tab:spans}, it can be concluded that simply adding the location embedding to the segment features yields the optimal span representations in our span refining decoder. The reason for why the \textbf{add} operation outperforms \textbf{cat-fn} may stem from the adverse effects of including an extra FC layer on the structure of the encoded multi-modal features. The comparison between the \textbf{add} operation and the \textbf{feature} approach highlights the importance of incorporating the span location information.

\begin{table}[h]
    \centering
    \resizebox{.85\linewidth}{!}{
    \scriptsize
    \renewcommand\arraystretch{0.85}
    \begin{tabular}{l|ccc}
      \toprule[1pt]
      Scaling & R1@0.3 & R1@0.5 & R1@0.7 \\ 
      \midrule[1pt]
        feature  & \textbf{76.71} & 61.43 & 38.73 \\ 
        cat-fn  & 75.03 & 61.35 & 37.82 \\ 
        add & 76.53 & \textbf{62.30} & \textbf{40.05} \\ 
      \bottomrule[1pt]
  \end{tabular}}
  \caption{\textbf{Span proposal representations}. Impact of the representations of the noisy spans in our span refining decoder.}
  \label{tab:spans}
\end{table}

\begin{figure}[h]
\centering
\includegraphics[width=8.4cm]{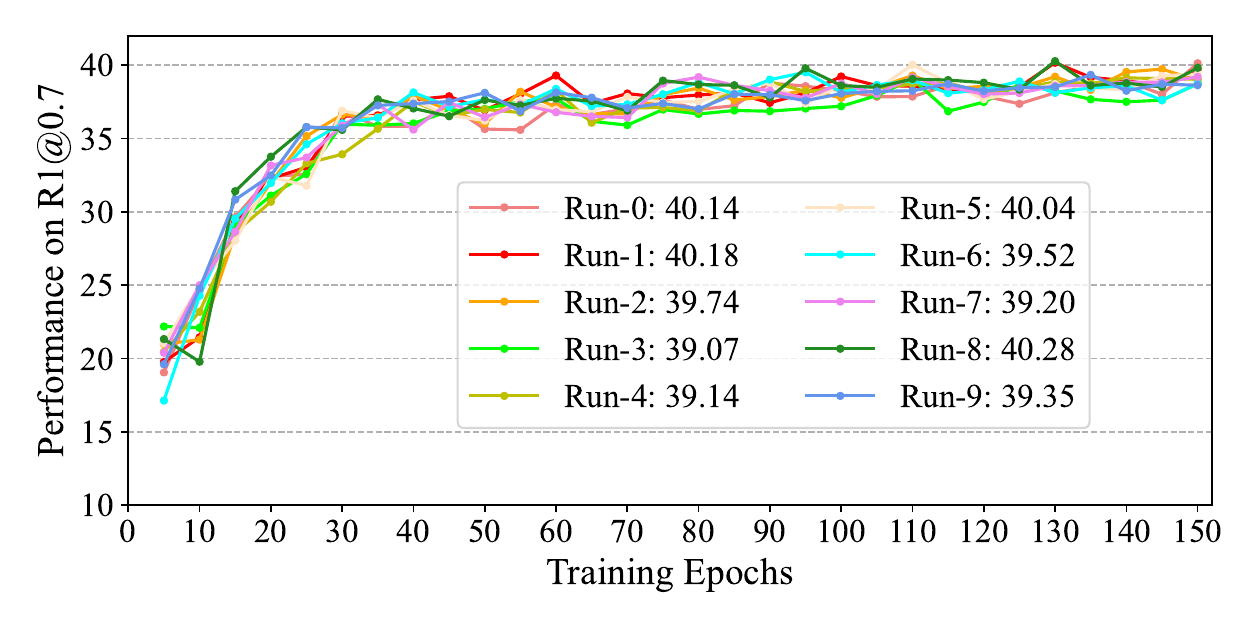}
\caption{\textbf{Stability}. All 10 runs are conducted on Charades-STA dataset. This result demonstrates the stability of DiffusionVG under different random seeds.}
\label{fig:stability}
\end{figure}

\noindent \textbf{5. Stability and Compatibility}.
Given that DiffusionVG generates predicted target spans from Gaussian noise inputs in inference and adds sampled noises to the ground-truth span during training, it is worth exploring whether its performance heavily depends on the sampled noises, namely, the random seeds. To demonstrate its stability under different random seeds, we repeat the training process of our DiffusionVG on Charades-STA~\cite{Gao_Sun_Yang_Nevatia_2017} dataset 10 times with distinct random seeds. We showcase all the validation results on R1@0.7 through the training process in Fig.~\ref{fig:stability}. Specifically, the best result and the worst result among the 10 runs are 39.07 and 40.28 in R1@0.7, respectively, while the standard deviation for all runs remains low at 0.44.

To study the compatibility between varying numbers of queries used during training and inference in DiffusionVG, we perform experiments utilizing 1, 5, and 10 queries during training, and evaluate each of these models with 1, 5, and 10 queries in inference. The results are summarized in Table.~\ref{tab:compatibility}. Notably, regardless of the number of queries employed during training, a consistent performance improvement is observed across all models as the evaluation queries increases from 1 to 5. However, as the number of queries in evaluation exceeds 5, the performance either experiences a slight decline or remains static. Since the effects of the number of evaluation queries are independent of the number of queries used in training, we can conclude that different query quantities employed in training and inference are mutually compatible. The sole influence of the number of training queries on model performance emerges when training queries are insufficient (first line in Table.~\ref{tab:compatibility}), leading to limited data augmentation and hindered convergence for the model.

\begin{table}[h]
    \centering
    \renewcommand\arraystretch{1.2}
    \begin{tabular}{l|ccc}
      \toprule[1pt]
      \diagbox{Train}{Test} & 1-query & 5-query & 10-query \\ 
      \hline
        1-query   & 38.13 & \textbf{38.78} &  38.60   \\ 
        5-query   & 38.80 & \textbf{40.05} &  39.91 \\ 
        10-query  & 38.54 & 39.97 &  \textbf{40.02} \\ 
      \bottomrule[1pt]
  \end{tabular}
  \caption{\textbf{Compatibility}. Ablation study on the compatibility between the varying numbers of queries utilized in training and inference stages in DiffusionVG.}
  \label{tab:compatibility}
\end{table}

\begin{table}[h]
    \centering
    \renewcommand\arraystretch{1}
    \begin{tabular}{l|ccc}
      \toprule[1pt]
      Voting & R1@0.3 & R1@0.5 & R1@0.7 \\
      \midrule[1pt]
        w/o  & 76.11 & 61.32 & 38.19 \\ 
        with   & \textbf{76.53} & \textbf{62.30} & \textbf{40.05} \\ 

      \bottomrule[1pt]
  \end{tabular}
  \caption{\textbf{Voting strategy}. Impact of the post voting process on DiffusionVG.}
  \label{tab:voting}
\end{table}

\noindent \textbf{6. Voting Strategy}.
To validate the effectiveness of our voting strategy, we employ a counterpart mechanism wherein the final predicted target span is randomly chosen from all predictions. As shown in Table.~\ref{tab:voting}, with our voting strategy to determine the best predicted target span, the performance of DiffusionVG gains a satisfactory improvement of 1.86\% in R1@0.7. Even though all queries are targeted at recovering the same target span, it is important to consider that each prediction may exhibit a certain level of deviation. Our voting strategy aims to identify the common area among all predictions, while simultaneously mitigating their imprecise boundaries, thus results in the optimal localization of the target span.

\subsection*{D. Implementation Details}
\label{sec:implement}
We train DiffusionVG using AdamW~\cite{Loshchilov_Hutter_2017} optimizer with a learning rate of $1e^{-4}$ and a weight decay of $1e^{-4}$. The batch size is set to 64 on all of the three datasets. For a fair comparison with existing VG methods, we extract video features using both I3D~\cite{Carreira_Zisserman_2017} and VGG-Net~\cite{simonyan2014very} for Charades-STA, while employing C3D~\cite{Tran_Bourdev_Fergus_Torresani_Paluri_2015} for ActivityNet Captions and TACoS. For both Charades-STA and Activitynet Captions, we set the length of the video features $K$ to 72, while the length is set to 128 for TACoS. We extract the sentence features with a pretrained DistilBERT-base model~\cite{sanh2019distilbert} from the \textit{Hugging face Transformers}~\footnote[1]{\url{https://huggingface.co/}} library. We exhibit the parameters in our Transformer model (video-centered encoder, span refining decoder and the DETR-like baselines of our implementation) in Table.~\ref{tab:transformer}. For the diffusion strategy, we adopt a monotonically decreasing cosine schedule to control the noises and set the diffusion steps $T=1000$. We empirically set the balance weights as $\lambda_{\ell 1}=1.5$, $\lambda_{giou}=1$, while the scaling factor $\lambda$ is set to 2. 


\begin{table}[h]
    \centering
    \resizebox{1.0 \linewidth}{!}{
    \begin{tabular}{p{80pt}|p{140pt}}
      \toprule[1pt]
      Parameter & Value  \\ 
      \midrule[1pt]
        Encoder layers  & 4 \\  
        Decoder layers  & 2 \\ 
        Attention heads & 8 \\ 
        Dropout & 0.1 \\
        Hidden dimension  & 256 \\ 
        FFN dimension  & 1024 \\
        Activation & ReLU \\
      \bottomrule[1pt]
  \end{tabular}}
  \caption{Transformer paremeters.}
  \label{tab:transformer}
\end{table}

\subsection*{E. More Visualizations}
\begin{figure*}[t]
\centering
\includegraphics[width=1.0\linewidth]{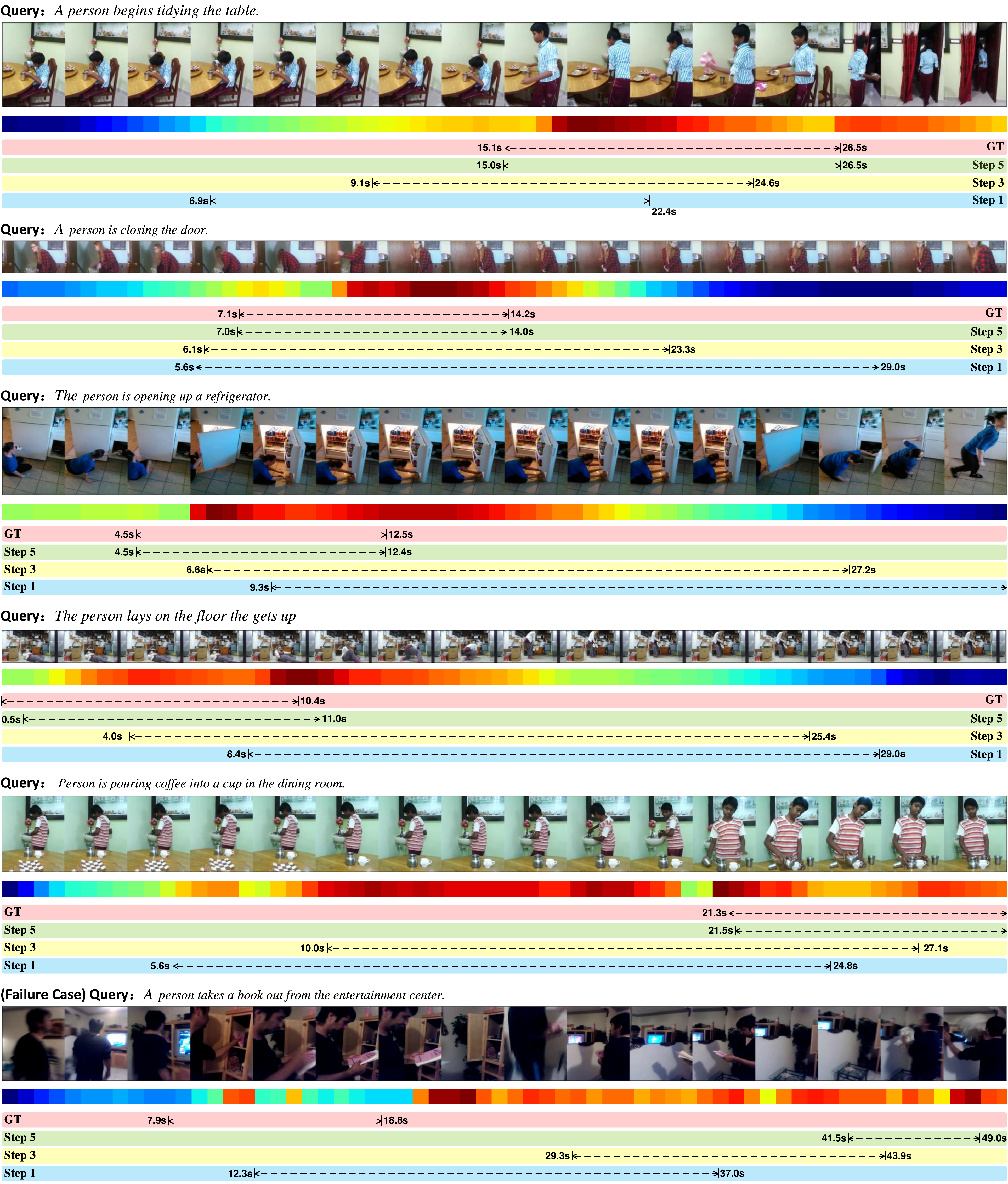}
\vspace{-0.4cm}
\caption{More visualized examples of our proposed DiffusionVG on Charades-STA test set. Additionally, we visualize the sentence-to-video attention map in each sample, where the average attention values for all words in a sentence are computed to yield the sentence-video level attentions.}
\vspace{-0.4cm}
\label{fig:more_vis}
\end{figure*}

We present more qualitative results on the test set of Charades-STA in Fig.~\ref{fig:more_vis} to further analyze the specific mechanisms in our proposed DiffusionVG. Additionally, we visualize the sentence-to-video attention map from the last cross-attention layer of our video-centered multi-modal encoder. Based on these qualitative results, we summarized several noteworthy observations: 

\noindent (1) Our video-centered multi-modal encoder can precisely capture the intricate correlations between the sentence query and video representations, with the highest attention predominantly focused on the corresponding target moment within the video in most cases (1, 2, 3, 4).

\noindent (2) The encoded multi-modal information is of vital importance since it provides indispensable guidance and serves as conditions for generating the target span through the denoising diffusion process. As shown in our failure case (6), the inaccurate attentions of the sentence query may result in deviations during the generation of the target span.

\noindent (3) Despite the significance of the accurate sentence-to-video attention, in cases where the error in the sentence-to-video attention remains within an acceptable range of deviation from the target moment, our DiffusionVG can effectively rectify the errors in attention within the denoising diffusion process, ultimately enabling precise span localization. As illustrate in case 5, the sentence-to-query attention from our video-centered encoder emphasizes on the middle part of the video, whereas the target moment is actually situated in the end of the video. However, by leveraging the inherent iterative refinement mechanisms of diffusion models, our DiffusionVG successfully generates the precise target moment. This observation further demonstrates the effectiveness of the iterative refining process in our DiffusionVG.

\end{document}